\documentclass[conference]{IEEEtran}

\usepackage{cite}
\ifCLASSINFOpdf
\usepackage{amsmath}
\usepackage{tikz}

\usepackage{url}
\usepackage{svg}
\usepackage{amsmath}
\usepackage{amssymb}

\usepackage{acronym}
\acrodef{DL}[DL]{Deep Learning}
\acrodef{NN}[NN]{Neural Network}
\acrodef{NNs}[NNs]{Neural Networks}
\acrodef{DNN}[DNN]{Deep Neural Network}
\acrodefplural{DNNs}[DNN]{Deep Neural Networks}
\acrodef{ML}[ML]{Machine Learning}
\acrodef{LSB}[LSB]{Least Significant Bit}
\acrodefplural{LSB}[LSBs]{Least Significant Bits}
\acrodef{HDL}[HDL]{Hardware Description Language}
\acrodef{PE}[PE]{Processing Element}
\acrodefplural{PE}[PEs]{Processing Elements}
\acrodef{axc}[AxC]{Approximate Computing}
\acrodef{SA}[SA]{Systolic Array}
\acrodefplural{SA}[SAs]{Systolic Arrays}
\acrodef{URE}[URE]{Uniform Recurrent Equations}
\acrodef{RIA}[RIA]{Regular Iterative Algorithm}
\acrodefplural{RIA}[RIAs]{Regular Iterative Algorithms}
\acrodef{HPC}[HPC]{High Performance Computing}
\acrodef{SEU}[SEU]{Single Event Upset}
\acrodefplural{SEU}[SEUs]{Single Event Upsets}
\acrodef{MAC}[MAC]{Multiply and ACcumulate}
\acrodef{FI}[FI]{Fault Injection}
\acrodef{SDC}[SDC]{Silent Data Corruption}
\acrodef{BER}[BER]{Bit Error Rate}
\acrodef{FIT}[FIT]{Failures In Time}

\usepackage{pgfplotstable}
\pgfmathdeclarefunction{gauss}{2}{%
  \pgfmathparse{1/(#2*sqrt(2*pi))*exp(-((x-#1)^2)/(2*#2^2))}%
}
\pgfplotsset{compat=1.11,
    /pgfplots/ybar legend/.style={
    /pgfplots/legend image code/.code={%
       \draw[##1,/tikz/.cd,yshift=-0.25em]
        (0cm,0cm) rectangle (3pt,0.8em);},
   },
}
\usepackage{colortbl}
\usepgfplotslibrary{colormaps}
    \def\addlegendimage{\csname pgfplots@addlegendimage\endcsname}

\pgfplotsset{ 
cycle list={%
{draw=black,mark=star,solid},
{draw=black, mark=square,solid}}}
\usepackage{breqn}

\usepackage[a4paper, total={185mm,239mm}]{geometry}

\def\BibTeX{{\rm B\kern-.05em{\sc i\kern-.025em b}\kern-.08em
    T\kern-.1667em\lower.7ex\hbox{E}\kern-.125emX}}

\usepackage{caption}
\usepackage{comment}
\usepackage{authblk}
\usepackage{amssymb}
\usepackage{fancyhdr}
\fancypagestyle{firstpage}
{
    \fancyhead[L]{\footnotesize © 2024 IEEE.  Personal use of this material is permitted.  Permission from IEEE must be obtained for all other uses, in any current or future media, including reprinting/republishing this material for advertising or promotional purposes, creating new collective works, for resale or redistribution to servers or lists, or reuse of any copyrighted component of this work in other works. This paper is accepted at the 27th International Symposium on Design and Diagnostics of Electronic Circuits and Systems (DDECS) 2024.}
    \fancyhead[R]{}
}
\usepackage{subcaption}
\IEEEoverridecommandlockouts \IEEEpubid{\makebox[\columnwidth]{979-8-3503-5934-3/24/\$31.00~\copyright{}2024 IEEE \hfill} \hspace{\columnsep}\makebox[\columnwidth]{ }}
\begin{document}
	
\title{SAFFIRA: 
	a Framework for Assessing the Reliability of Systolic-Array-Based DNN Accelerators
	}

\author[1*]{Mahdi Taheri}
\author[3,1]{Masoud Daneshtalab}
\author[1]{Jaan Raik}
\author[1]{Maksim Jenihhin}
\author[2*]{\\Salvatore Pappalardo}
\author[2]{Paul Jimenez}
\author[2]{Bastien Deveautour}
\author[2]{Alberto Bosio}

\affil[1]{Tallinn University of Technology, Tallinn, Estonia}
\affil[2]{Ecole Centrale de Lyon, Lyon, France}
\affil[3]{Mälardalen University, Västerås, Sweden}

\maketitle
\thispagestyle{firstpage}

\begin{abstract}     Systolic array has emerged as a prominent architecture for Deep Neural Network (DNN) hardware accelerators, providing high-throughput and low-latency performance essential for deploying DNNs across diverse applications. However, when used in safety-critical applications,  reliability assessment is mandatory to guarantee the correct behavior of DNN accelerators. While fault injection stands out as a well-established practical and robust method for reliability assessment, it is still a very time-consuming process. This paper addresses the time efficiency issue by introducing a novel hierarchical software-based hardware-aware fault injection strategy tailored for systolic array-based DNN accelerators. The uniform Recurrent Equations system is used for software modeling of the systolic-array core of the DNN accelerators. The approach demonstrates a reduction of the fault injection time up to 3$\times$ compared to the state-of-the-art hybrid (software/hardware) hardware-aware fault injection frameworks and more than 2000$\times$  compared to RT-level fault injection frameworks — without compromising accuracy. Additionally, we propose and evaluate a new reliability metric through experimental assessment. The performance of the framework is studied on state-of-the-art DNN benchmarks. 

\end{abstract}
\let\thefootnote\relax\footnote{* These authors contributed equally}
\let\svthefootnote\thefootnote


\addtocounter{footnote}{-1}\let\thefootnote\svthefootnote

\begin{IEEEkeywords}
hardware accelerator, systolic array, deep neural networks, fault simulation, reliability, resilience assessment
\end{IEEEkeywords}

\section{Introduction}

Assessing the reliability of a \ac{DNN} is not a trivial task: it depends on several factors, such as the training set, the data type, and the quality of the test set \cite{taheri2022dnn}. On top of that, we need to consider the hardware that performs the computations~\cite{ahmadilivani2023systematic} since specific platforms have specific faults~\cite{bosio2021emerging}.

\begin{figure}[t]
    \includegraphics[width=0.5\textwidth]{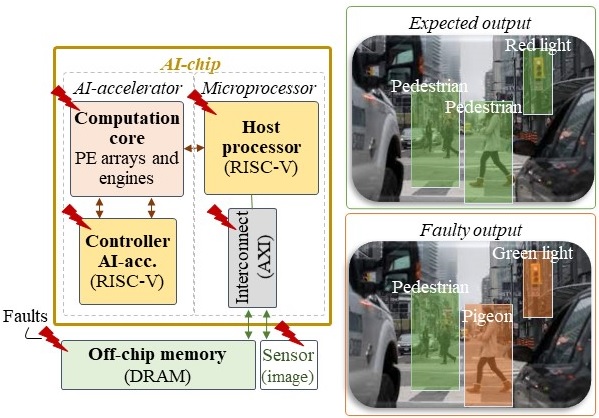}
    \centering
    \caption {DNN accelerator hardware reliability threats\cite{taheri2024exploration}}
    \label{fig:rel_thr}
\end{figure}

Many studies showed that hardware faults can greatly reduce the effectiveness of \acp{DNN} \cite{taheri2023appraiser}. As a result, there is a surge in research efforts to evaluate and enhance the reliability of DNNs. 
An example of faulty hardware is given in Figure \ref{fig:rel_thr}. This figure presents the possible fault locations in a DNN inference engine. This example shows the necessity of reliability assessment of DNNs. However, assessing DNN reliability is a challenging task \cite{taheri2023deepaxe}.

There are three main methodologies on DNNs' reliability assessment as irradiation-based, platform-based and simulation-based \cite{ruospo_2023_survey} in which 
simulation-based \ac{FI} is less expensive (in terms of equipment) and thus is also the most used in the research community \cite{ahmadilivani2023systematic}. The other advantages of FI are the possibility to model different fault scenarios precisely to assess their impact on DNNs without the need for extensive hardware resources and design time, and full control over network parameters and architecture.

On the other hand, DNN hardware-accelerator simulation for FI is computationally expensive and typically demands a substantial amount of time to complete a single inference \cite{ahmadilivani2023special}. This paper introduces a novel simulation flow and \ac{FI} tailored to significantly accelerate the injection process on systolic-array-based DNN hardware accelerators. The systolic-array core of the DNN accelerators is modeled using the Uniform Recurrent Equations (URE) system. The proposed injection flow has been implemented as an open-source tool named \textbf{SAFFIRA}, which stands for \textbf{Systolic Array simulator Framework for Fault Injection based Reliability Assessment}.
Simulation-based \ac{FI} is usually done either without considering the underlying hardware or through RTL (Register-Transfer Level) simulations known for their resource-intensive computations and time-consuming nature.  SAFFIRA is based on a \ac{SA} simulator, thus offering the advantage of being more precise than a hardware-agnostic tool, but much faster than traditional RTL-level simulations. Experimental results show a reduction of the fault injection time up to 3$\times$ compared to the state-of-the-art hybrid (software/hardware) hardware-aware fault injection frameworks and up to 2000$\times$  compared to RT-level fault injection frameworks — without compromising accuracy.



The key contributions of this paper are the following:

\begin{itemize}
\item introducing a hierarchical \textbf{methodology} for the hardware-accurate reliability assessment of \ac{SA} using a novel simulation-based fault injection approach by modeling the systolic-arrays using Uniform Recurrent Equations (URE) system;

\item presenting an \textbf{open-source tool} implementing the aforementioned methodology; 
\item introducing a new \textbf{metric} called \textbf{faulty distance} for reliability assessments of DNNs;
\item evaluating the performance of the framework on state-of-the-art DNN benchmarks



\end{itemize}

The rest of this paper is organized as follows. Section \ref{sec:related_works} presents the related works. Section \ref{sec:proposed_method} presents the proposed fault injection flow for \ac{SA}. Section \ref{sec:experiments} shows the experimental setup and results. Section \ref{sec:conclusions} concludes the paper.

\section{Related Works}
\label{sec:related_works}
This section discusses previous works targeting \acp{DNN} reliability assessment by using simulation-based \ac{FI}.

\subsection{Hardware-Agnostic FI Tools} 

Tools in this category perform fault injection without taking into account the underlying hardware. 
Some of these are capable of performing \ac{FI} directly in the \acp{DNN} models.
In this category, PyTorchFI \cite{mahmoud2020pytorchfi} and TensorFI \cite{narayanan2022fault} can inject faults into \ac{DNN} models respectively implemented in PyTorch, Tensorflow, and Keras.
All of these open-source frameworks can inject both permanent and transient faults into weights as well as activations given specific error rates such that it is possible to evaluate the accuracy loss. 

Moreover, to further enhance the efficiency, additional \ac{FI} tools have been introduced. For example, BinFI \cite{chen2019binfi} is an extension of TensorFI that aims at identifying critical bits in \acp{DNN}. Another tool, namely LLTFI \cite{agarwal2022lltfi}, is able to inject transient faults into specific instructions of DNN models in either PyTorch or TensorFlow. 

\subsection{Hardware-Aware FI Tools}
These tools can perform \ac{FI} in software, taking into account the relying hardware using some abstract models of the `DNN hardware accelerator. 


In \cite{pappalardo2023resilience}, the authors used an RTL model of a \ac{SA} to perform their experiments. 
Reference \cite{azizimazreah2018tolerating} maps a \ac{DNN} 
into the RTL implementation of the accelerator. They study the effect of transient faults in memory and datapath accurately. In these studies, \ac{FI} is performed in software while all of its parameters are integrated with the corresponding hardware components.
Authors in \cite{li2020soft} implemented their \ac{DNN} and the fault injector in software, inspired by an FPGA-based DNN accelerator. Moreover, in \cite{ozen2020low}, DNN and FI are implemented in Keras, and the architecture of a \ac{SA} accelerator is considered for a fault-tolerant design. Similarly, authors in \cite{jasemi2020enhancing} evaluate their proposed reliability improvement technique on memories in TensorFlow while injecting transient faults into the weights. PyTorch is used in \cite{ozen2020boosting} to implement the DNN, and transient faults are injected into activations (datapath or MAC units) and weights (memory) regarding the \ac{SA} accelerator model. Reference \cite{goldstein2021lightweight} also uses PyTorch and injects faults by a custom framework called TorchFI to inject faults into the outputs of CONV and FC layers of the network.

The effect of permanent faults at PEs' outputs is studied in \cite{burel2022mozart+} where the model of the accelerator is adopted from implementing the DNN in an N2D2 framework \cite{N2D2_fr}. Furthermore, authors in \cite{hoang2021tre} use PyTorch and study permanent faults in MAC units of an accelerator while training to improve the reliability at inference. Authors in \cite{tsai2021evaluating} developed a Keras-based accelerator simulator to study the effect of permanent faults on the on-chip memory of accelerators by injecting permanent faults into activations and weights. Weight remapping strategy in memory to decrease the effect of permanent faults is evaluated in \cite{nguyen2021low} using Ares. SCALE-Sim \cite{samajdar2018scale}, a systolic CNN accelerator simulator, is adopted in \cite{zhao2022fsa} to study permanent faults in PEs and computing arrays in systolic array-based accelerators.

Similar to the Hardware-Independent platform, faults are injected based on \ac{BER}, or fault rate, and experiments are repeated to reach 95\% confidence level and 1\% error margin \cite{ozen2020low}. 
In general, the main drawbacks in the existing reliability assessment methods for \acp{DNN} can be summarized as follows:

\begin{itemize}
	\item There is no software FI framework in hardware-aware platforms. Hence, there is a potential for DNN accelerator simulators to be exploited or developed for the reliability assessment of DNNs;
	\item Several FI research works carry out accuracy loss and fault classification as an evaluation of reliability. Also, some works considered FIT (Failure In Time) \cite{li2017understanding}. However, there is still an urgent need to present DNN-specific metrics for reliability evaluation. In this work, we are introducing a new metric called faulty distance to provide a better understanding of the network resilience.
\end{itemize}

\section{Proposed Methodology}
\label{sec:proposed_method}
\begin{figure}
    \centering
    \includegraphics[width = 0.5\textwidth]{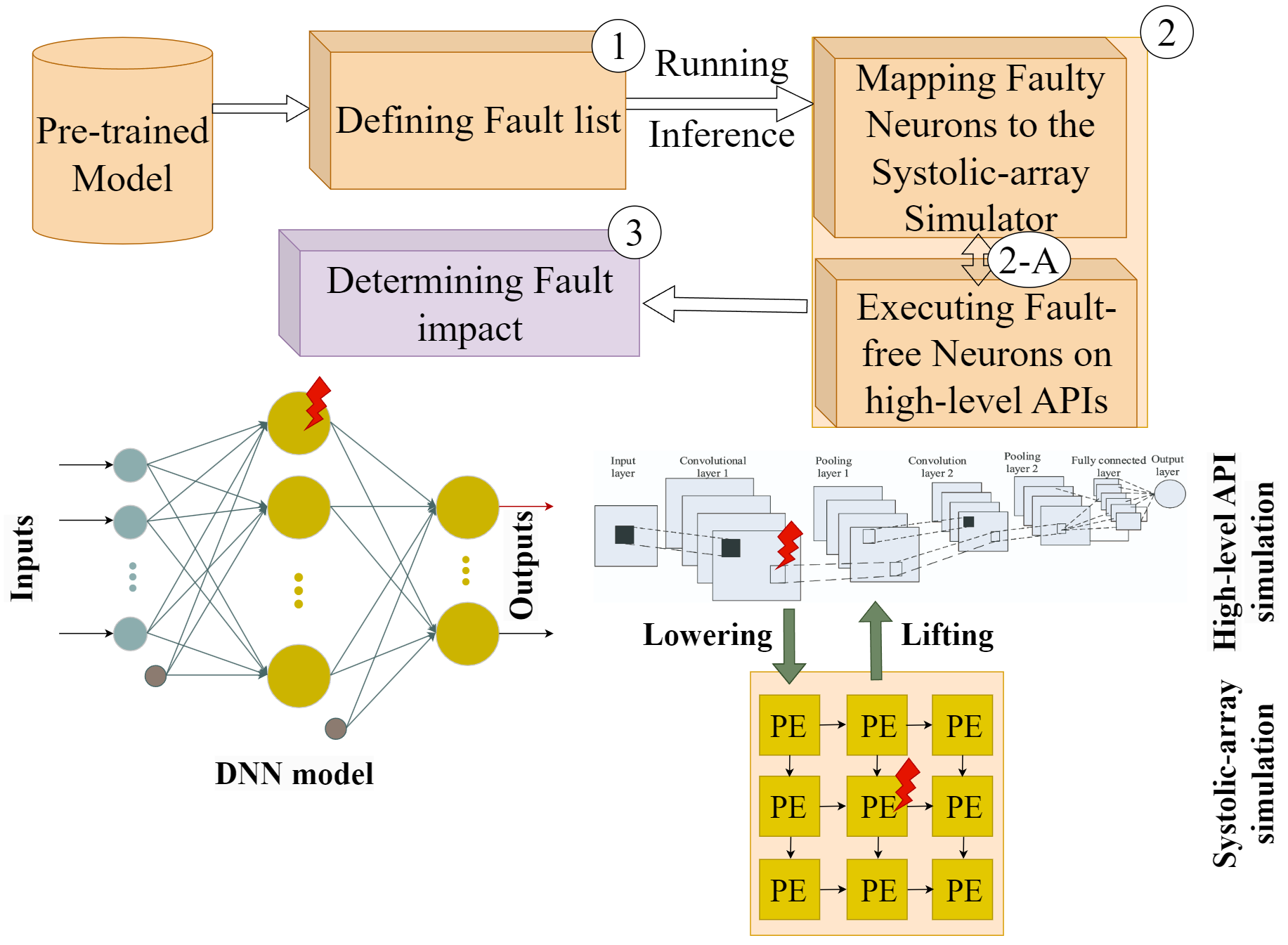}
    \caption{SAFFIRA methodology}
    \label{method}
\end{figure}

The proposed methodology for the SAFFIRA framework is illustrated in Fig. \ref{method}. After providing the trained network parameters and architecture, in step one, the fault list is generated. Possible fault locations can be defined by the user or can be a random fault list generated based on the network parameters by the framework. Faults can be selected as transient or permanent faults targeting different activations of the DNN. Then, in step two, the fault injection campaign is performed at the systolic-array simulation environment in Python, and the rest of the network is executed at the high-level API (e.g.  Pytorch) to speed up the process. In this step, switching between high-level API and systolic-array simulator (2-A) is done by a method called LoLif, which is described in subsection \ref{subsec:convolutions}. Finally, the reliability of the network and the impact of the faults are reported at step three by different metrics.

SAFFIRA supports various data representations, including fixed-point, integer, and floating-point formats. This framework also supports various relevant mapping to systolic-array architecture scenarios (e.g. output stationary, weight stationary, etc.). These flexibilities allow researchers to adapt the framework to different applications and tailor the reliability assessment to specific hardware requirements.


\subsection{Hardware simulations}

SAFFIRA is a \ac{SA} model based on the \ac{URE} system. 
As described by \cite{quinton1984automatic}, it is possible to generate a \ac{SA} that solves the problem described by a \ac{URE} system.
In the case of SAFFIRA, the URE system is the one associated with matrix multiplication since it is the operation deployed on the \ac{SA} for DNN execution. The following subsection presents the formal details needed to perform a simulation followed by performing fault injection in such a context, and finally, the strategies strictly related to \acp{DNN} are covered.
	
	\subsubsection{Mathematical formalism}
	
	A \ac{URE} system is defined on top of an integer lattice $L_n$ of points $p$ in the n-dimensional Euclidean space $E_n$. The goal is to solve a system of equations associated with the variables $x_1(p), x_2(p), \cdots, x_m(p)$ for all points $p \in R$, where $R \subseteq L_n$ \cite{quinton1984automatic}.
	This system can be either uni-variate or multi-variate. Here, only uni-variate case is considered, thus the system would have the following form.
	
	\begin{align*}
		&x_1(p) = f[x_1(p-w_1), \cdots, x_m(p-w_p)], \\
		&x_2(p) = x_2(p-w_2), \\
		&\phantom{100000}\vdots \\
		&x_m(p) = x_m(p - w_p).
	\end{align*}
	
	The points $p - w_{i_k}$ belong to $L_n$. The vectors $w_k$ are constants independent of $p$ and this is why they are said to have \textit{uniform dependence}.
	Each equation $x_i(p)$ depends on the points $p - w_{i_k}$. 
	
	The authors of \cite{quinton1984automatic} showed a strategy to model a \ac{SA} starting from the problem to solve. Specifically, the authors explain three steps:
	
	\begin{enumerate}
		\item find a \ac{URE} system for the problem to solve,
		\item find a timing function compatible with the dependencies of the \ac{URE} system,
		\item find an allocation function to map the \ac{URE} onto a finite architecture.
	\end{enumerate}
	

The main idea is to project the space $E_n$ twice: the first time, the resulting points will correspond to the spatial arrangement of each \ac{PE}. The second projection determines iso-temporal planes, identifying operations that are computed during the same clock cycle but on different \acp{PE}; each plane corresponds to a different clock cycle. The space-projection matrix $P$ and the temporal dimension vector $\pi$ are used later.

\subsubsection{Convolutions}
\label{subsec:convolutions}
The strategy explained above opens the possibility to implement a variety of algorithms as a systolic array.
Based on the literature, it is possible to perform a convolution as a matrix multiplication \cite{hadjis2015caffe}. The experiments shown below are performed using a systolic array to perform the matrix multiplication $C = A \times B$. The associated \ac{URE} is the following. 

\begin{align*}
    c(i, j, k) = c(i, j, &k-1) + a(i, j-1, k) \times b(i-1, j, k) \\
    a(i, j, k) &= a(i, j-1, k) \\
    b(i, j, k) &= b(i-1, j, k)\\
    \text{initial conditions}&\\
    a(i,0,k) &= a_{ik}, \quad \forall i, k\\
    b(0, j, k) &= b_{kj}, \quad \forall k, j\\
    c(i, j, 0) &= 0, \quad \forall i, j\\
\end{align*}
where $p = (i, j, k) \in R \subseteq L_n$, in which $i$, $j$ and $k$ assume values between $1$ and $N1$, $N2$, $N3$ respectively. $N1$, $N2$ and $N3$ are problem parameters such that $A \in \mathbb{R}^{N_1, N_3}, B \in \mathbb{R}^{N_3, N_2}, C \in \mathbb{R}^{N_1, N_2}$.

\begin{figure}
    \centering
    \includegraphics[width=0.72\columnwidth]{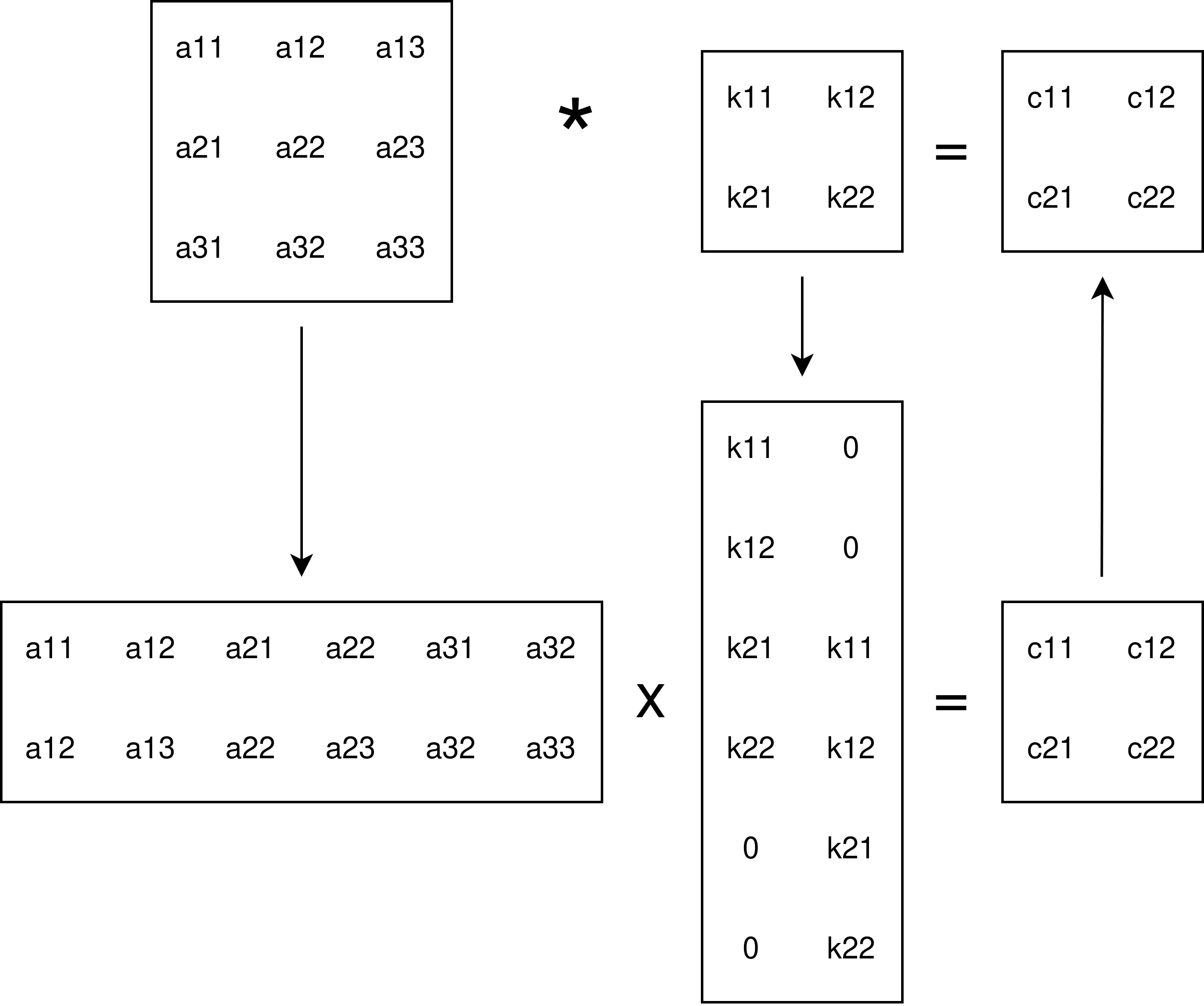}
    \caption{LoLif example. Applied transformations are similar to im2col and im2row.}
    \label{lolifexample}
\end{figure}

When it comes to performing a convolution, the input matrices must be \textit{reshaped} such that the result of the \ac{SA} is a convolution. In this paper, this concept is called LoLif, which stands for Lowering and Lifting strategies. This idea is explained in \cite{hadjis2015caffe}. If computing a convolution $C = A * B$ is needed, it can be implemented as a transformation $lif$ of the matrix multiplication of transformed matrices $low_a(A) \times low_b(B)$. In formulas: 

$$
C = lif( low_a(A) \times low_b(B) ),
$$
where $Lif$, $Low_a$ and $Low_b$ are corresponding transformations, as shown in the example of Fig. \ref{lolifexample} 

\subsubsection{Simulation and Injections}

In order to perform the simulation, it is sufficient to solve the system shown above. Nevertheless, this method gives the possibility of injecting faults in the values in a hardware-aware fashion. To achieve the injection, it is sufficient to change the values $a(p)$, $b(p)$, $c(p)$ for specific points $p$. The faulty values must then be propagated to the following \acp{PE}.
Given that each point $p$ is projected to the physical space $r = (x, y)$ using the physical space-projection matrix, $r = P p$, it can be inferred that how the injected values are propagated through the different \acp{PE}.
Specifically, for some dependence vector $d$ for the different labeled variables $a, b, c$. Looking at the system above, the following can be observed: $d_a = (0, 1, 0)$, $d_b = (1, 0, 0)$ and $d_c = (0, 0, 1)$. Afterwards, the propagation direction can be found using the same relationship shown before: $\delta x_i = P d_i, i = \{a,b,c\}$. This means that the value of $a$ in some \ac{PE} in position $s$ will be propagated to the \ac{PE} in position $s + \delta x_a$. The same reasoning can be done for the time, supposing that a fault is propagated not only in space but also in time. We can compute $\delta t_i = \pi d_i$. For simplicity, $\pi = (1,1,1)$ is fixed to reduce the exploration space. In this case, the time dependency $\delta t_i$ will always be $1$: $\delta t_i = 1$. 

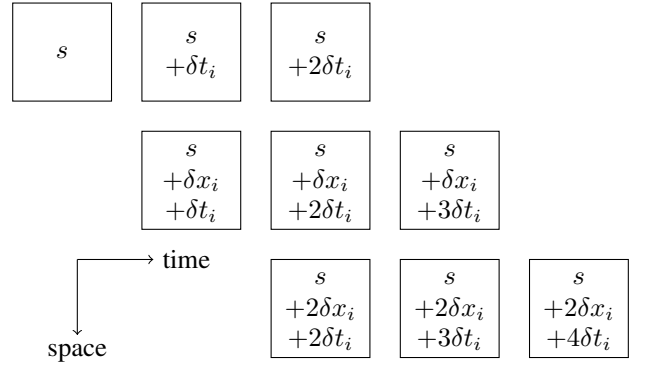
\begin{figure}[tb]
    \centering
    \begin{tikzpicture}
        \newcommand\scale{1.7}
        \newcommand\boxsize{1.3}
        \foreach \i in {1,...,3}{	
            \foreach \j in {1,...,3}{
                \draw (\scale * \i + \scale*\j, -\scale*\j) rectangle (\scale*\i+\scale*\j+\boxsize, -\scale*\j-\boxsize);
            }
        }
        
        \newcommand\s{$s$}
        \newcommand\dt{\delta t_{i}}
        \newcommand\dx{\delta x_{i}}
        
        \draw (2 * \scale + \boxsize/2, -\scale-\boxsize/2) node {\s};
        
        \draw (3 * \scale + \boxsize/2, -\scale-\boxsize/2+0.2) node {\s};
        \draw (3*\scale + \boxsize/2, -\scale-\boxsize/2-0.2) node {$+ \dt$};
        
        \draw (4*\scale + \boxsize/2, -\scale-\boxsize/2+0.2) node {\s};
        \draw (4*\scale + \boxsize/2, -\scale-\boxsize/2-0.2) node {$+ 2 \dt$};
        
        \draw (3*\scale+\boxsize/2, -2*\scale-\boxsize/2+0.4) node {\s};
        \draw (3*\scale+\boxsize/2, -2*\scale-\boxsize/2) node {$+ \dx$};
        \draw (3*\scale+\boxsize/2, -2*\scale-\boxsize/2-0.4) node {$+\dt$};
        
        \draw (4*\scale+\boxsize/2, -2*\scale-\boxsize/2+0.4) node {\s};
        \draw (4*\scale+\boxsize/2, -2*\scale-\boxsize/2) node {$+ \dx$};
        \draw (4*\scale+\boxsize/2, -2*\scale-\boxsize/2-0.4) node {$+ 2 \dt$};
        
        \draw (5*\scale+\boxsize/2, -2*\scale-\boxsize/2+0.4) node {\s};
        \draw (5*\scale+\boxsize/2, -2*\scale-\boxsize/2) node {$+ \dx$};
        \draw (5*\scale+\boxsize/2, -2*\scale-\boxsize/2-0.4) node {$ + 3 \dt$};
        
        \draw (4*\scale+\boxsize/2, -3*\scale-\boxsize/2+0.4) node {\s};
        \draw (4*\scale+\boxsize/2, -3*\scale-\boxsize/2) node {$+ 2 \dx$};
        \draw (4*\scale+\boxsize/2, -3*\scale-\boxsize/2-0.4) node {$ + 2 \dt$};
        
        \draw (5*\scale+\boxsize/2, -3*\scale-\boxsize/2+0.4) node {\s};
        \draw (5*\scale+\boxsize/2, -3*\scale-\boxsize/2) node {$+ 2 \dx$};
        \draw (5*\scale+\boxsize/2, -3*\scale-\boxsize/2-0.4) node {$ + 3 \dt$};
        
        \draw (6*\scale+\boxsize/2, -3*\scale-\boxsize/2+0.4) node {\s};
        \draw (6*\scale+\boxsize/2, -3*\scale-\boxsize/2) node {$+ 2 \dx$};
        \draw (6*\scale+\boxsize/2, -3*\scale-\boxsize/2-0.4) node {$ + 4 \dt$};
        
        \draw[->] (2.5*\scale, -3*\scale) -- (2.5*\scale+1, -3*\scale) node[at end, right] {time};
        \draw[->] (2.5*\scale, -3*\scale) -- (2.5*\scale, -3*\scale-1) node[at end, below]{space};
    \end{tikzpicture}
    \caption{When injecting element $s$, the fault is propagated in time (thus affecting elements $s + \delta t_i$ and $s + 2 \delta t_i$) and in space (forwarding the faulty value to neighboring elements $s + \delta x_i + \delta t_i$, $s + 2\delta x_i + \delta t_i$ and so on).}
    \label{fig:injected_elements}
\end{figure}

Figure \ref{fig:injected_elements} shows an example. In this case, an injection in the element $s = (x, y, t)$ on the generic line $i$ is done between times $0$ and $\infty$.
The injected elements are visible in the figure.
Specifically, the fault will propagate in time, thus injecting also $s + \delta t_i$ and $s + 2 \delta t_i$.
In the same way, this fault will propagate in space, to the element cascading from $s$. Note that the value propagation only happens after each clock cycle. This means that the next injected element will be displaced also in time, thus injecting element $s + \delta x_i + \delta t_i$. In the same way, the latter will propagate to the following element on the following clock cycle, thus injecting element $s + 2 \delta x_i + 2 \delta t_i$ and so on.

The set of points belonging to the injection can be transposed back into the iteration space $E_n$ using the pseudo-inverse $P^{-1}$. The set of points identified with this strategy will be subject to injection.
Formally, injection is as a function $h$ applied to a variable:

$$
a(p) = h(a(p-w_a))
$$

\section{Experiments and Results}
\label{sec:experiments}

Two different sets of experiments are performed using SAFFIRA.
First, a fault injection based on the permanent-fault model is performed on two different quantization versions of the LeNet-5 network (8-bit and 16-bits integers). The second set of experience is performing fault injection based on the transient fault model in the three different benchmarks (AlexNet, VGG-16 and ResNet-18). All networks are fully quantized to INT data type, including all activations, weights, and biases. The base accuracies are reported in the table \ref{baseaccuracy}
\begin{table}[]
\caption{Base accuracy of networks under test}
\label{baseaccuracy}
\centering
\begin{tabular}{|c|c|}
\hline
DNN                    & accuracy (\%)\\ \hline
8-bit LeNet-5 (MNIST)  & 93.8                          \\ \hline
16-bit LeNet-5 (MNIST) & 95.4                          \\ \hline
AlexNet (CIFAR-10)     & 78.0                          \\ \hline
VGG-16 (CIFAR-10)      & 93.4                          \\ \hline
ResNet-18 (CIFAR-10)   & 93.8                          \\ \hline
\end{tabular}
\end{table}

The \ac{SA} model for these experiments is output stationary. This means that its physical-space projection matrix $P$ is as follows:

$$
P = \begin{pmatrix}
	1 & 0 & 0 \\
	0 & 1 & 0 \\
\end{pmatrix}
$$

Such a matrix corresponds to a rectangular \ac{SA} with $N1 \times N2$ \acp{PE}. Please note that with this projection, the variable $c$ (i.e. the partial sum) is a \textit{stationary variable} since it is always available on the same \ac{PE} regardless of the iteration. Whether a variable is stationary or not depends on the employed projection.

In all experiments, fault injection is repeated several times to reach an acceptable confidence level, based on \cite{leveugle2009statistical}. This work provides an equation to reach 95\% confidence level and 1\% error margin.

\subsection{Fault Classification}

The \ac{DNN} resilience is evaluated by comparing the output probability vector of the golden run (i.e. the \ac{DNN} that behaves as expected, without faults) and the faulty run (i.e. the \ac{DNN} that includes the fault).
The \ac{SDC} rate is defined as the proportion of faults that caused misclassification in comparison with the golden model \cite{9926241}. 

In addition, the targeted hardware reliability can be calculated by differentiating \ac{SDC} rates of injected transient faults into defined classes and calculating \ac{FIT} for the accelerator (\textit{accel}) by its components (\textit{comp}) with \eqref{eq:FIT1} in which \(FIT_{raw}\) is provided by the manufacturer, \(Size_{comp}\) is the total number of the component bits, and \(SDC_{comp}\) is obtained by FI.
\begin{equation*}
	FIT_{accel} = \sum_{comp} FIT_{raw} \times Size_{comp} \times SDC_{comp} 
	\label{eq:FIT1}
\end{equation*}

Finally, \textbf{faulty distance} is proposed. This metric can be used to evaluate the resilience of classifications \acp{DNN}.
Supposing the golden probability vector is $G$, the faulty probability vector is $F$ and the function $ag(\cdot)$ corresponds to the $\text{argmax}$ function, then the faulty distance function $d_f$ is defined as follows.

$$
d_f = (1 - \frac{G \cdot F}{ ||G|| \cdot ||F||}) \cdot (ag(F) - ag(G))
$$

In this metric, cosine similarity is being used $cos \theta = \frac{G \cdot F}{ ||G|| \cdot ||F||}$. Cosine similarity serves as a metric for assessing the resemblance between two non-zero vectors within an inner product space. Representing the cosine of the angle between the vectors, this measure calculates similarity by normalizing their dot product. In our study, we utilize cosine similarity to evaluate the entirety of generated probabilities across various classes in both faulty and golden modes.
The cosine similarity metric yields values within the range of -1 to 1. Proximity to 1 signifies a high degree of similarity between vectors.
Therefore, the faulty distance metric gives 0 when the faulty output corresponds to the correct classification. The bigger the metric, the worse the misclassification is.

\subsection{Results}

Table \ref{tab:metrics} shows the results of the \ac{FI} for permanent fault injection experiments on LeNet-5 with the different metrics. It can be seen that this network was highly susceptible to the injected permanent faults. Specifically, the SDC-1 and SDC-5 are very high: on average, about 82\% of the time, the faulty inference misclassified the input; furthermore, about 93.5\% of the inputs were completely missed since the correct label was below the fifth position. 
The SDC-10\% and SDC-20\% rates are very high as well: more than 95\% of the inputs had the correct class with a probability much too low than expected. Average Faulty Distance (AFD) is also reported that shows the 16-bit network in this particular case, is more reliable compared to the 8-bit network in the presence of permanent faults in the systolic architecture.

These results show that the \ac{DNN} used was not usable in a safety-critical environment. This result was expected since the network was not trained to withstand stuck-at faults like the ones injected.

\begin{table}[h]
\centering
\caption{\ac{FI} experiments results on two LeNet-5}
\label{tab:metrics}
\begin{tabular}{lcc}
\hline
 Metric                   &  16bit  & 8bit\\
\hline
 SDC-1 (\%)               & 77.84   & 87.70 \\
 SDC-5  (\%)              & 93.05   & 94.49 \\
 FIT (failures/$10^9$ hours)                      &  4.9e-4 & 5.0e-4  \\
 SDC-10\% (\%)              & 98.16   & 98.53 \\
 SDC-20\% (\%)              & 96.21   & 96.97 \\
 AFD	& -0.04	  &  -0.53 \\
\hline
\end{tabular}
\end{table}
For the second experiment, only SDC and AFD are reported in table \ref{benchmark}.

\begin{table*}[]
\centering
\caption{Reliability analysis of different state-of-the-art DNN benchmarks}
\label{benchmark}
\begin{tabular}{|c|c|c|c|c|c|}
\hline
DNN                  & SDC-1 & SDC-5 & SDC-10\% & SDC-20\% & AVF      \\ \hline
AlexNet (CIFAR-10)   & 4.3   & 29.1  & 13.1     & 9.7      & $7.1\times 10^{-2}$ \\ \hline
VGG-16 (CIFAR-10)    & 3.0   & 40.0  & 46.5     & 84.5     & $1.9\times 10^{-3}$ \\ \hline
ResNet-18 (CIFAR-10) & 1.5   & 23.0  & 16.5     & 82       & $1.6\times 10^{-3}$ \\ \hline
\end{tabular}
\end{table*}

\begin{figure}[h]
	\centering
	\begin{subfigure}{0.65\columnwidth}
		\includegraphics[width=\textwidth]{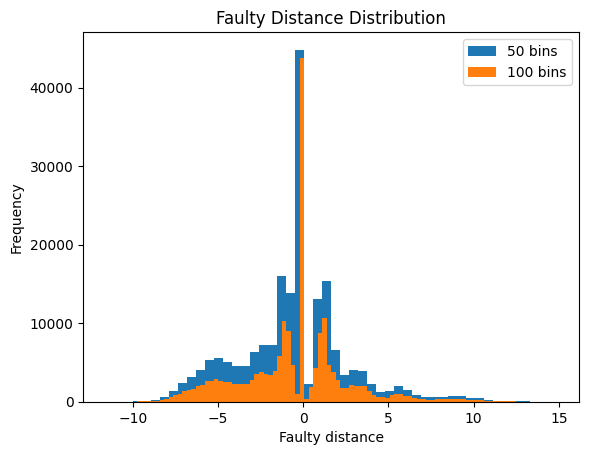}
		\caption{Faulty Distance on the 8bit network}
		\label{fig:fd_plot_8bit}
	\end{subfigure}
	\begin{subfigure}{0.65\columnwidth}
		\includegraphics[width=\textwidth]{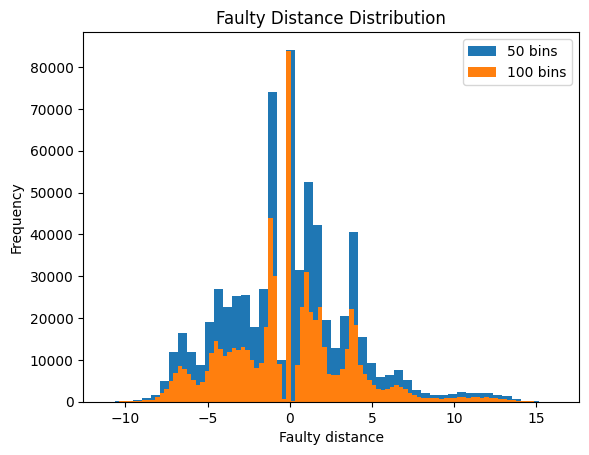}
		\caption{Faulty Distance on the 16bit network}
		\label{fig:fd_plot_16bit}
	\end{subfigure}
	\caption{Histogram plot of the Faulty distance values}
	\label{fig:faulty_distance_plots}
\end{figure}

\subsection{Faulty Distance}
In the previous subsection, only the average faulty distance was shown. Nevertheless, this metric can be looked through with more details when plotted as a histogram. Figure \ref{fig:faulty_distance_plots} shows the histograms (both with 50 and 100 bins) of the metrics per each experiment. It is possible to see a peak at 0, which corresponds to all the correctly classified inputs. The height of that column is precisely the same as the complement of the SDC-1 metric. On top of that, it is possible to see two different, yet similar, trends for the two networks. Figure \ref{fig:fd_plot_8bit} shows other three peaks: around $+1$, $-1$ and $-5$. This means that, although most of the inputs were mis-classified, the difference with the golden vector was not extremely big, in general.
On the other hand, figure \ref{fig:fd_plot_16bit} shows many more peaks, this means that it is more difficult to predict how a fault will propagate in this case.

\subsection{Computation Time}
The experiments were performed on a server using python3 with an Intel Xeon Silver 4210, with a total number of 40 cores.
SAFFIRA completes 500 inferences of two convolutional layers, with the same systolic array, in about 10 minutes with minimal optimization. This means a total of about 16.3 simulations per second. For comparison, by utilizing the framework presented in \cite{pappalardo2023fault} to perform fault injection on the same networks as this work, on average, 5.8 simulations per second are executed. The mentioned framework is the state-of-the-art hybrid (software/hardware codesign) hardware-aware fault injection framework. Therefore, SAFFIRA provides about 2.8$\times$ speed up by performing the same analysis. Also, the same fault injection campaign is performed at the RT level using QuestaSim. The results show 0.007 simulation per second, which is 2100$\times$ slower than the proposed method in this work.
	
\section{Conclusions}
\label{sec:conclusions}
This paper presents a novel hierarchical fault injection strategy for systolic arrays, addressing the time efficiency issue by introducing a novel hierarchical software-based hardware-aware fault injection strategy tailored for systolic array-based DNN implementations. The approach demonstrates a reduction of the fault injection time up to threefold compared to the state-of-the-art hybrid (software/hardware) hardware-aware fault injection frameworks and more than 2000$\times$ compared to RT-level fault injection frameworks — without compromising accuracy. Additionally, we propose and evaluate a new reliability metric through experimental assessment. The performance of the framework is studied on state-of-the-art DNN benchmarks.

\section{Acknowledgement}
This work was supported in part by the Estonian Research Council grant PUT PRG1467 "CRASHLESS“ and by Estonian-French PARROT project "EnTrustED".
	
\bibliographystyle{ieeetr}
\bibliography{refs}

\end{document}